\documentclass{article}
\usepackage{lss4absa}

\usepackage{times}
\usepackage{soul}
\usepackage{url}
\usepackage[hidelinks]{hyperref}
\usepackage[utf8]{inputenc}
\usepackage[small]{caption}
\usepackage{graphicx}
\usepackage{amsmath}
\usepackage{amsthm}
\usepackage{booktabs}
\usepackage{algorithm}
\usepackage{algorithmic}

\usepackage{amssymb}
\usepackage{bm}
\usepackage{amstext}
\usepackage{makecell}
\usepackage{multirow}
\usepackage{booktabs}
\usepackage{bbding}
\usepackage{color}
\usepackage{arydshln}

\title{Learn from Structural Scope: Improving Aspect-Level Sentiment Analysis with Hybrid Graph Convolutional Networks}

\author{
Lvxiaowei Xu$^1$
\and
Xiaoxuan Pang$^2$\and
Jianwang Wu$^{1}$\and
Ming Cai $^{1}$\footnote{ Corresponding author}\and
Jiawei Peng$^1$
\affiliations
$^1$Department of Computer Science and Technology, Zhejiang University\\
$^2$Alibaba Group-China Digital Commerce-Tao Cai Cai-Tech Dept-Data Intelligence Growth\\
\emails
\{xlxw, wujw, cm, pengjw\}@zju.edu.cn,
pangxiaoxuan.pxx@alibaba-inc.com
}

\begin{document}

\maketitle

\begin{abstract}
Aspect-level sentiment analysis aims to determine the sentiment polarity towards a specific target in a sentence. The main challenge of this task is to effectively model the relation between targets and sentiments so as to filter out noisy opinion words from irrelevant targets. Most recent efforts capture relations through target-sentiment pairs or opinion spans from a word-level or phrase-level perspective. Based on the observation that targets and sentiments essentially establish relations following the grammatical hierarchy of phrase-clause-sentence structure, it is hopeful to exploit comprehensive syntactic information for better guiding the learning process. Therefore, we introduce the concept of \emph{Scope}, which outlines a structural text region related to a specific target. To jointly learn structural \emph{Scope} and predict the sentiment polarity, we propose a hybrid graph convolutional network (HGCN) to synthesize information from constituency tree and dependency tree, exploring the potential of linking two syntax parsing methods to enrich the representation. Experimental results on four public datasets illustrate that our HGCN model outperforms current state-of-the-art baselines.
\end{abstract}

\section{Introduction}
\begin{figure}[t]
	\centering
	\includegraphics[width=0.9\columnwidth]{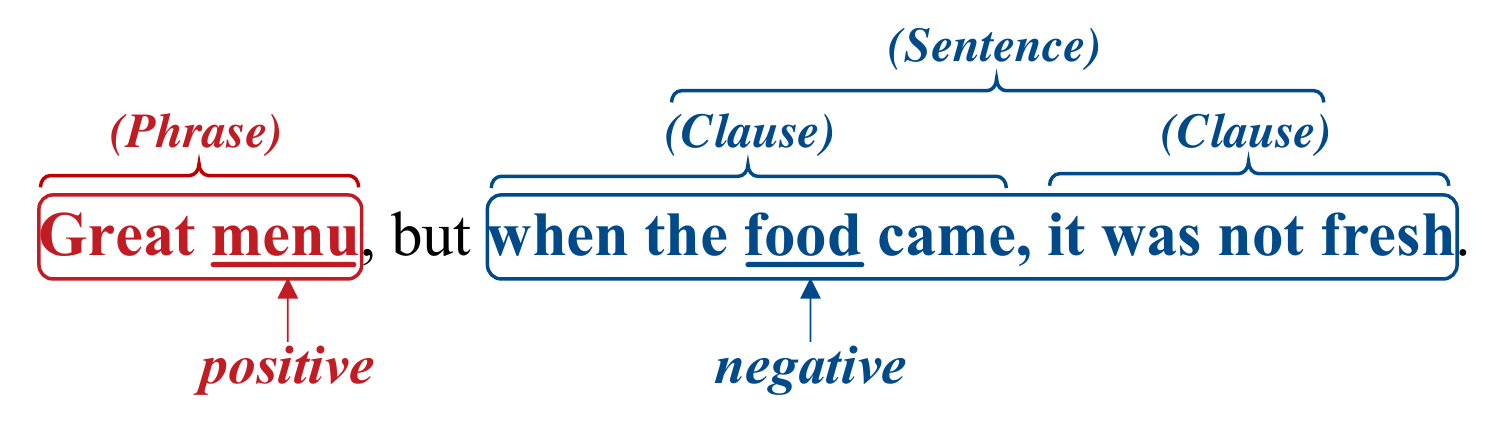} 
	\caption{An instance of a restaurant review with two target terms which have opposite sentiments. The \emph{Scopes} are highlighted with a box and contain their corresponding target.}
	\label{fig1}
\end{figure}

Aspect-level sentiment analysis (ALSA) is a fine-grained classification task, aiming at identifying opinion polarities towards specific entities called targets. Figure \ref{fig1} shows a review commenting on the restaurant with two target terms ``\emph{menu}'' and ``\emph{food}''. The sentiment polarities over them are positive and negative respectively. Since multiple targets may appear in one sentence and convey different sentiments, the main challenge of ALSA task is to effectively filter out noisy or misleading opinion words from irrelevant targets and extract distraction-free text from opinion expressions accurately.

Considerable efforts have been devoted to overcoming the difficulties. Their works can be broadly divided into the following three categories. \textbf{Attention-based models:} Attention mechanism has been widely adopted to learn target-specific features \cite{wang2016attention,ma2017interactive} so as to help models concentrate on crucial parts of a sentence for sentiment classification. However, the attention scatters across the whole sentence and is prone to attend on noisy opinion words from other targets. \textbf{Syntax-based models:} Syntax-based method aims to model syntactic structures that exist in the sentence. Among them, dependency tree \cite{zhang2019aspect,sun2019aspect,wang2020relational} has been mostly used to capture the structural dependencies between targets and the sentiment expressions. Nevertheless, multi-hop dependency relations between target and unrelated opinion words that are susceptible to confusing and misleading the model. \textbf{Span-based models:} Span-based models \cite{wang2018learning,hu2019learning,xu2020aspect} apply a divide-and-conquer strategy to handle the ALSA task. Specifically, the model extracts an opinion span towards each target based on a soft or hard selection approach, and then focuses on the phrase-level features for sentiment classification. Therefore, it is difficult to generate structurally complete spans for complicated sentences due to lack of syntactic information.

To tackle aforementioned problems, we introduce a concept of structural \emph{Scope}, which is a well-structured and continuous text region expressing target-specific opinion. \emph{Scope} delineates the exact context of each target so as to alleviate the distraction problem caused by opinion words from other targets. Compared with opinion span, \emph{Scope} is characterized by a well-defined structure, which highlights the learning of various syntactic features (e.g., phrase, clause and sentence) related to the target to ensure structural and semantic completeness. Figure \ref{fig1} demonstrates an example containing two \emph{Scopes} with their corresponding targets in different structures. It combines a phrase-level relation (a noun phrase connecting the target ``\emph{menu}’’ and opinion word ``\emph{great}’’) and a sentence-level relation (the target ``\emph{food}’’ linking the remote opinion words ``\emph{not fresh}’’ outside of the local clause through a coreference relation). This demonstrates that the targets and sentiments establish relations following the grammatical hierarchy of phrase-clause-sentence structure, which substantially filters out noisy opinion words from irrelevant targets.

Three observations motivate us to aggregate syntactic information of constituency tree and dependency tree to learn structural \emph{Scope}. First, constituency parsing decomposes a sentence into phrases and combines them hierarchically into clauses and sentences, which naturally follows the phrase-clause-sentence structure of \emph{Scope} and can serve as the foundation to encode structural information. Second, dependency parsing is capable of capturing long-range dependencies such as coreference relation, which can help the model obtain the sentence-level connection. Third, constituency parsing displays the entire sentence structure and relations, while dependency parsing explores relations between words. They present two complementary sides of syntactic relations in sentences. Therefore, it would be beneficial to incorporate constituency tree  and dependency tree into the model to learn more comprehensive syntactic relations.

On the basis of above analysis, we present a hybrid graph convolutional network (HGCN) to jointly learn the structural \emph{Scope} and determine the sentiment polarity. For encoding the constituency tree, we propose the constituency GCN module and the computation is performed in three stages. First, the representations of the constituents are composed by the representations of the words and the embeddings of constituent labels. Second, graph convolutional networks (GCNs) \cite{kipf2016semi} are exploited to learn constituent representations. Third, a constituent-token attention mechanism is applied to decompose back the information to word representations. Regarding the dependency tree, we utilize a relational graph convolutional network (RGCN) to obtain syntactic representation with labeled edges. Finally, we incorporate such heterogeneous information to determine the \emph{Scope} region through the conditional random field (CRF) \cite{lafferty2001conditional}, and to identify the opinion polarities as well.

The main contributions of this work include:
\begin{itemize}
\item We introduce the concept of \emph{Scope} as a syntactically-informed method to capture the relations between targets and sentiments, which could generate distraction-free opinion expressions for sentiment classification.
\item We propose a constituency GCN module equipped with constituent-token attention mechanism to encode structural representations from constituency tree. 
\item We present a new HGCN model to synthesize constituency relations and dependency relations, exploiting the complementary strengths of two syntax parsing methods to learn syntactic relations for \emph{Scope}.
\end{itemize}

\section{Model}
\subsection{Definition of \emph{Scope}}
To eliminate noisy opinion words from unrelated targets, we propose a syntactic structure guided modeling concept called \emph{Scope}. In general, \emph{Scope} contains a specific target and its dependent sentiment expression within a continuous and minimum text in a sentence. Specifically, we first define a constituent\footnote{Note that we regard non-terminal node above the part-of-speech (POS) level as constituent whereas the terminal node is the word.} candidate set $C_s$ to hold a group of constituents (denoted as $c_i$), which include a consecutive text span from target to opinion words in their leaf nodes (denoted as $Node(c_i)$).  Each $c_i$ can be categorized into three levels according to grammatical definitions, which are phrase level (e.g., VP, NP), clause level (e.g., SBAR) and sentence level (e.g., S). \emph{Scope} is then defined as seeking the span of text corresponding to the constituent with the fewest leaf nodes in $C_s$:
\begin{equation}
Node(C_s\left[\operatorname{argmin}(\left[\operatorname{count}(Node(c_i)), c_i \in C_s\right])\right]) \label{eq0}
\end{equation}
where $\operatorname{count}(\cdot)$ calculates the number of nodes excluding extraneous components (e.g., adjunct and punctuation).

\subsection{Graph Convolutional Network}

We leverage graph convolutional networks over constituency tree and dependency tree to encode structural information. GCNs are neural networks that compute the representation of a node conditioned on its neighboring nodes in a given graph. Each node is updated by aggregating the propagated message from the neighboring nodes through multilayer GCNs. Given a graph $\mathcal{G}$ contains sets of nodes $\mathcal{V}$ and edges $\mathcal{E}$. Following the idea of \cite{kipf2016semi}, we allow $\mathcal{G}$ to have self-loops. After that, we can obtain the adjacency matrix $\mathcal{A} \in \mathbb{R}^{n \times n}$ according to the $\mathcal{G}$, where $n$ denotes the size of $\mathcal{V}$. The $A_{ij}$ in the adjacency matrix reflects the connection between the node $i$ and node $j$. Specifically, $A_{ij} = 1$ if node $i$ is connected to node $j$ and in the other scenario $Aij = 0$. Then GCN is able to propagate information over the paths and update node representations. In general, stacking $l$ layers of GCN allows aggregation of information across the $l$-th order neighborhood. In such an operation, the representation of each node is updated with normalization factor as follow:
\begin{equation}
	h_{i}^{(l+1)}= \sigma \left(\sum_{j=1}^{n} c_{i} A_{i j}\left(W^{(l)} h_{j}^{(l)}+b^{(l)}\right)\right) \label{eq1}
\end{equation}
where $h_{j}^{(l)}$ is the hidden representation for node $j$ that is generated by $l$-th layer of the GCN. $W^{(l)}$ and $b^{(l)}$ are trainable parameters of weights and bias in $l$-th layer, respectively. $c_i$ is the normalization factor calculated as $c_i = 1 / d_i$ and $d_i$ denotes the degree of node $i$ in the graph, which is computed as $d_{i}=\sum_{j=1}^{k} A_{i j} $. And $\sigma$ is a non-linear activation function, (e.g., $\text{RELU} (\cdot)$ \cite{glorot2011deep}).

\subsection{Hybrid Graph Convolutional Network}

\begin{figure*}[t]
	\centering
	\includegraphics[width=1.\textwidth]{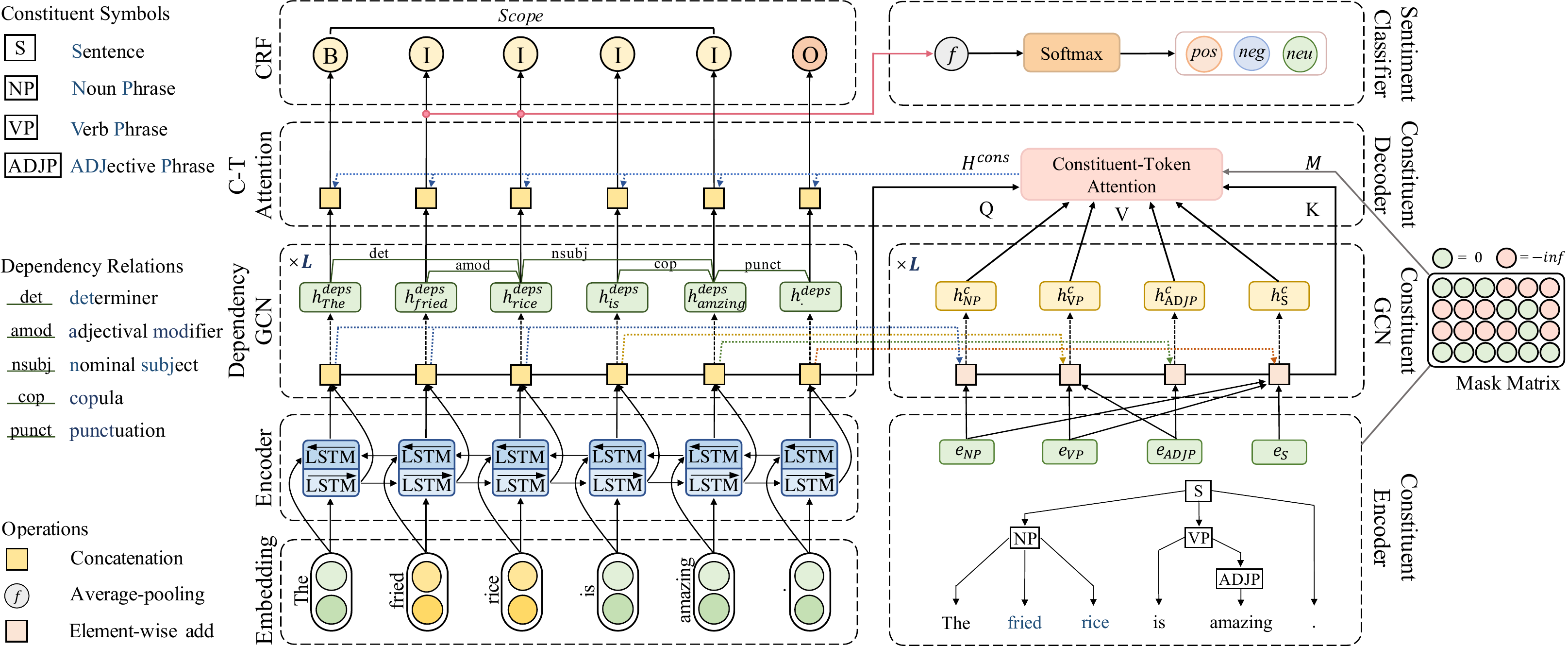} 
	\caption{Overview of hybrid graph convolutional network (HGCN).}
	\label{fig2}
\end{figure*}

Our model aims to extract the \emph{Scope} relevant to the target term and predict the sentiment polarity jointly. Figure \ref{fig2} gives an overview of our HGCN model. In the aspect-level sentiment classification task, a sentence with $n$ words $\left\{w_{1}, \ldots, w_{i}, w_{i+1}, \ldots, w_{j}, w_{j+1}, \ldots, w_{n}\right\}$ is given, where the target $\left\{w_{i}, \ldots, w_{j}\right\}$ is a sub-sequence of the sentence.

Then, BiLSTM \cite{graves2013speech} or BERT \cite{devlin2019bert} is applied as sentence encoder to integrate contextual information into hidden representations of words. For the BiLSTM encoder, we first embed each word into low-dimensional vector space by looking up an embedding matrix $E_{emb} \in \mathbb{R}^{d \times |V|}$, where $d$ denotes the dimension of word embedding and $|V|$ is the size of vocabulary. Then we feed the embedding of words into BiLSTM encoder for obtaining contextual hidden representations $H = \{h_1, h_2, \cdots, h_n\}$, where $h_i \in \mathbb{R}^{2d}$ is the hidden representations from BiLSTM. As for the BERT, we construct an aspect oriented context ``[CLS] sentence [SEP] aspect [SEP]’’ as input to obtain contextual hidden representations of the sentence. Moreover, there is a gap between word-piece representations of BERT and word-level based syntactic features. Therefore, we compose the representations of the sub-word as word representations with average-pooling to tackle this mismatch. Then the contextual representations are fed into the Constituency GCN (CGCN) and Dependency GCN (DGCN), respectively. 

Eventually, we concatenate representations from DGCN and CGCN module to form syntactic representations. After that, a CRF layer is utilized to generate BIO-tags for modeling the \emph{Scope} with syntactic representations and we aggregate the syntactic representations over the target terms via pooling to determine sentiment polarity jointly. In the following, we will elaborate the details of our proposed HGCN model.

\subsubsection{Constituency GCN (CGCN)}

The constituency tree breaks a sentence into phrases, and then they can be composed hierarchically into clause and sentence level from bottom to top. Intuitively, it is reasonable to utilize constituency tree to encode structural information for modeling \emph{Scope}. In order to incorporate such structural constituency syntax, we propose CGCN module with multi-stage architecture for obtaining constituency-aware representations. Compared with the SpanGCN model \cite{marcheggiani2020graph}, we design the constituent-token (C-T) attention to reassign the representation of words as a substitute for BiLSTM, which reduces the computational complexity and attains better performance. As shown in Figure \ref{fig2}, the architecture of CGCN is composed by three sub-modules: constituent encoder, constituent GCN and constituent decoder.

\paragraph{Constituent Encoder.}

The constituency tree is composed of constituents ($\mathcal{V}_{c}$) and words ($\mathcal{V}_{w}$). In order to obtain representations of constituents, we first embed each constituent label using $E_{cons} \in \mathbb{R}^{d_c \times |\mathcal{N}|}$, where $d_c$ is the dimension of embedding and $|\mathcal{N}|$ denotes the number of constituents. Then we assign the representation of each constituent $v \in \mathcal{V}$ in this stage relying on the embedding of the constituent label (e.g., NP or VP) and their child nodes set $\mathcal{S}_i \subseteq (\mathcal{V}_{c} \cup \mathcal{V}_{w}) $. Take Figure \ref{fig2} as an example. The constituent ``S’’ has three child nodes. Two of them belong to $\mathcal{V}_{c}$ and the rest one belongs to $\mathcal{V}_{w}$. Its representation is obtained through averaging the embeddings of its own label, the label embedding of its two child constituents and the representations of word ``.''. 

\paragraph{Constituent GCN.}

In this stage, it enables the interactive information passing from each constituent and ensures information of the child nodes, which can be integrated into the representation of their immediate parent node and vice versa. GCN is operated on the graph constructed by constituency tree where the nodes represent all constituents in set $\mathcal{V}_{c}$ and the edges correspond to the connection between constituents and their child nodes in both directions and each constituent is connected to itself on account of the self-loop. After that, structural syntactic representation $H^{\text{c}} = \left\{h_{1}^{\text {c }}, h_{2}^{\text {c }}, \ldots, h_{m}^{\text {c }}\right\}$ can be obtained from Equation \ref{eq1} with layer normalization applied, where $m$ refers to the number of constituents.

\paragraph{Constituent Decoder.}
After the message passing in GCN module, structural syntactic information are injected into the representation of constituents. At this point, we propose the constituent-token (C-T) attention mechanism to assign the information back to each word. In our implementation, the attention weights indicate the contribution of the word to the representation of corresponding constituent which can be computed in constituent encoder stage. Then the allocation coefficient (i.e. attention scores) can be computed as below:
\begin{equation}
	A^{cons} = \operatorname{softmax}\left(\frac{QW_q \times \left(KW_k\right)^{\mathrm{T}}}{\sqrt{d}} + M\right) \label{eq2}
\end{equation}
where the matrices $Q$ and $K$ are the representation of words and the representation of constituents composed by constituent encoder, respectively. $W_q$ and $W_k$ denote the trainable parameters, while $d$ is the dimensionality of the hidden representations. In addition, $M$ stands for a mask matrix to ensure that information can only be injected back to the words from their parent constituent. Here, the dot product is adopted to calculate relatedness between words and constituents.

Then we can obtain constituency representations of words based on attention scores, which can be formulated as:
\begin{equation}
 	H^{\text{cons}} = A^{cons} H^{\text{c}} \label{eq3}
\end{equation}

After the above three-stage process, we obtain the constituency representations $H^{\text{cons}} = \left\{h_{1}^{\text {cons}}, h_{2}^{\text {cons}}, \ldots, h_{n}^{\text {cons}}\right\}$.

\subsubsection{Dependency GCN (DGCN)}

The DGCN is used to capture long-range dependencies such as coreference relationships. Intuitively, different dependency relations should have distinctive impacts on adjacent nodes. However, most of the previous work fails to consider the relationship of the edges. Inspired by \cite{schlichtkrull2018modeling}, we utilize relational graph convolutional network (RGCN) to encode dependency structure with labeled edges.

Firstly, in order to integrate the relationship information of the edges, we  embed each kind of edge label using a randomly initialized embedding matrix $E_{rel} \in \mathbb{R}^{d_r \times |\mathcal{M}|}$ where $d_r$ refers to the dimension of word embedding and $|\mathcal{M}|$ is the number of relation types. Then we can obtain the representation of the relation between node $i$ and node $j$ as $r_{ij}$. We consider the representation $r_{ij}$ as the gate-value to determine how much of the information will be updated to neighboring nodes. Thus the adjacency matrix $A_{ij}^{d}$ takes the form:
\begin{equation}
	A_{ij}^{d} = \sigma\left(r_{ij} W_r + b_r\right) \label{eq4}
\end{equation}
where $W_r$ and $b_r$ denote trainable parameters of weights and bias. $\sigma(\cdot)$ refers to the sigmoid function. After that, we regard the hidden representation $H$ from BiLSTM or BERT encoder as initial node representations in DGCN. After that, we can compute dependency graph representation $ H^{\text{deps}} = \{h_1^{\text{deps}}, h_2^{\text{deps}}, \cdots, h_n^{\text{deps}}\} $ based on Equation \ref{eq1}. 

Finally, we apply concatenation operation on CGCN and DGCN to obtain the syntactic representations $H^{syn}$ for \emph{Scope} modeling. Next an average pooling operation is utilized to generate the sentiment representation $H^{senti}$ for the ALSA task. Then $H^{senti} $ is fed into a linear layer and mapped to probabilities of different polarities by a softmax function, i.e.,
\begin{equation}
	p(y)=\operatorname{softmax}\left(W_{p}H^{senti} + b_{p}\right) \label{eq5}
\end{equation}
where $W_p$ and $b_p$ are the learned weights and bias.

\subsection{Training and Testing}

During the training process, we define a joint loss for selection of \emph{Scope} and sentiment classification after obtaining the syntactic representation from CGCN and DGCN:
\begin{equation}
	\mathcal{L}(\Theta)=  \mathcal{L}_{polarity} + \gamma \mathcal{L}_{scope} + \lambda\|\Theta\|_{2}  \label{eq6}
\end{equation}
where $\gamma$ is the coupling co-efficiency that regulates the two losses. $\Theta$ represents all trainable parameters in our proposed HGCN model and $\lambda$ refers to the coefficient of $L_2$-regularization. Moreover, $\mathcal{L}_{polarity}$ denotes a standard three-way cross-entropy loss defined for the ALSA task. $\mathcal{L}_{scope}$ is defined as the negative conditional log-likelihood loss of the CRF layer for \emph{Scope} modeling. We can formulate them as:
\begin{equation}
	\mathcal{L}_{polarity}=- \sum_{i} \hat{y}_{i} \log p\left(y_{i}\right) \label{eq7}
\end{equation}
\begin{equation}
	\mathcal{L}_{scope}=-\log P(\mathbf{y} \mid \mathbf{u}; W, b)  \label{eq8}
\end{equation}
where $\hat{y}_{i}$ denotes the ground truth, $i$ is the $i$-th sentiment polarity and $W$, $b$ are trainable parameters of weights and bias.

In addition, as for the \emph{Scope} selection task, Viterbi algorithm is utilized to predict the most likely label assignment of sequence during the testing process.

\begin{table}[t]	
	\centering
	%\fontsize{10}{11}\selectfont
	\begin{tabular}{ccccccccc}
		\toprule
		\multirow{2}{*}{\textbf{Dataset}}&
		\multicolumn{2}{c}{\textbf{Positive}}&\multicolumn{2}{c}{\textbf{Neutral}}&\multicolumn{2}{c}{\textbf{Negative}}\cr \cline{2-7}
		&\rule[0pt]{0pt}{10pt}\textbf{Train}&\textbf{Test}&\textbf{Train}&\textbf{Test}&\textbf{Train}&\textbf{Test}\cr
		\midrule
		%\hline
		Rest14       &2164&727&637&196&807&196  \cr
		Lap14        &976&337&455&167&851&128  \cr
		Rest15       &912&326&36&34&256&182  \cr
		Rest16       &1657&611&101&44&748&204  \cr
		\bottomrule
	\end{tabular}
	\caption{Statistics for the four experimental datasets.}
	\label{tab1}
\end{table}

\section{Experiments}
\subsection{Experimental Setups.}
\paragraph{Datasets.} We conduct experiments of aspect-level sentiment analysis task on four benchmark datasets. As the statistics of the datasets shown in Table 1, there are  restaurant-domain (Rest14, Rest15, Rest16) datasets and laptop-domain (Lap14) taken from SemEval \cite{maria2014semeval,pontiki2015semeval,pontiki2016semeval}. In order to utilize \emph{Scope}, we develop a semi-automated annotation tool and then ask 4 experienced annotators and 2 verifiers to annotate the datasets with BIO-tags of \emph{Scope}. It is worth mentioning that our tool can automatically complete the preliminary annotation based on syntax and designed rules. Then annotators make fine refinements on this basis and only 26.4\% of the samples require manual adjustments. There are more details in Appendix A.

\paragraph{Baselines.}
Baselines can be categorized into three groups.  $\bullet$ \textbf{1) Attention-based method.} \textbf{ATAE-LSTM} \cite{wang2016attention} uses aspect embedding and attention mechanism in ALSA tasks. \textbf{AOA} \cite{huang2018aspect} utilizes two LSTMs and an interactive attention mechanism to generate representations for the aspect and sentence. \textbf{MGAN} \cite{fan2018multi} proposes a multigrained attention network to capture word-level interactions between target term and context. $\bullet$ \textbf{2) Syntactic-based method.} \textbf{ASGCN} \cite{zhang2019aspect} and \textbf{CDT} \cite{sun2019aspect} first apply graph convolutional network for encoding aspect-specific word representations. \textbf{BiGCN} \cite{zhang2020convolution} adopts hierarchical graph structure to encode dependency and word co-occurrence information. \textbf{InterGCN} \cite{liang2020jointly} employs a GCN over a dependency tree to learn aspect representations. \textbf{RGAT}, \textbf{RGAT+BERT} 
\cite{wang2020relational} utilize a relational graph attention network to encode the pruned dependency trees. \textbf{DGEDT}, \textbf{DGEDT+BERT} \cite{tang2020dependency} provide a dependency graph enhanced dual-transformer network to encode heterogeneous information. \textbf{DualGCN}, \textbf{DualGCN+BERT} \cite{rui2021dual} incorporate syntactic structure and semantic relevance to generate features. $\bullet$ \textbf{3) Span-based method. SA-LSTM}  \cite{wang2018learning} presents a segmentation attention model to distill the sentiment semantics for capturing the structural dependencies with external opinion data. \textbf{MCRF-SA} \cite{xu2020aspect} proposes a hard selection approach to determine the opinions by self-critical reinforcement learning with annotated opinions.

\begin{table}[t]
	\centering
	\begin{tabular}{p{2.29cm} p{1.03cm}<{\centering} p{1.03cm}<{\centering} p{1.03cm}<{\centering}  p{1.03cm}<{\centering}}
		\toprule
	    \textbf{Model} & \textbf{Rest14} & \textbf{Lap14} & \textbf{Rest15} & \textbf{Rest16} \cr
		\midrule
		\multicolumn{5}{l}{$\bullet$ \textbf{Attention-based methods}} \cr
		%\hline
		ATAE-LSTM        &77.01&70.09&76.60&84.36\cr 
		AOA              &79.97&72.62&78.17&87.50\cr
		MGAN             &79.96&71.81&78.15&85.29\cr \hdashline
		\multicolumn{5}{l}{$\bullet$ \textbf{Syntactic-based methods}} \cr
		ASGCN            &80.80&74.61&79.43&88.02\cr
		CDT              &81.74&74.65&77.94&83.88\cr
		BiGCN            &80.83&74.96&80.33&88.47\cr
		InterGCN         &81.26&75.06&79.54&88.52\cr
		DGEDT            &80.89&73.04&78.32&86.52\cr
		RGAT             &81.50&73.72&78.78&88.21\cr 
		DualGCN          &82.85&76.66&80.16&88.19\cr \hdashline
		\multicolumn{5}{l}{$\bullet$ \textbf{Span-based methods}} \cr
		SA-LSTM          &77.64&70.67&76.63&85.96\cr 
		MCRF-SA          &80.94&74.11&78.71&87.76\cr \hdashline
		%\multicolumn{5}{l}{$\bullet$ \textbf{Ours methods}} \cr
		\makecell[l]{\textbf{HGCN} \\ \textbf{(Ours)}}\rule[0pt]{0pt}{15pt}   
		                 &\makecell[c]{\textbf{84.09} \\ \textbf{(82.91)}}\rule[0pt]{0pt}{15pt}
		                 &\makecell[c]{\textbf{78.64} \\ \textbf{(76.82)}}\rule[0pt]{0pt}{15pt}
		                 &\makecell[c]{\textbf{82.66} \\ \textbf{(80.81)}}\rule[0pt]{0pt}{15pt}
		                 &\makecell[c]{\textbf{89.84} \\ \textbf{(88.92)}}\rule[0pt]{0pt}{15pt}\cr \hline 
		\multicolumn{5}{l}{$\bullet$ \textbf{Models with BERT}} \cr 
		BERT+Finetune    &84.30&77.45&81.85&90.24\cr 
		DGEDT+BERT       &85.99&78.88&82.95&90.94\cr 
		RGAT+BERT        &85.91&79.10&83.15&91.39\cr 
		DualGCN+BERT     &86.29&79.51&83.78&91.43\cr \hdashline
		\makecell[l]{\textbf{HGCN+BERT} \\ \textbf{(Ours)}}\rule[0pt]{0pt}{15pt}   
		                 &\makecell[c]{\textbf{87.41} \\ \textbf{(86.45)}}\rule[0pt]{0pt}{15pt}
		                 &\makecell[c]{\textbf{81.49} \\ \textbf{(79.59)}}\rule[0pt]{0pt}{15pt}
		                 &\makecell[c]{\textbf{85.61} \\ \textbf{(83.91)}}\rule[0pt]{0pt}{15pt}
		                 &\makecell[c]{\textbf{93.02} \\ \textbf{(91.72)}}\rule[0pt]{0pt}{15pt}\cr 
	\bottomrule
	\end{tabular}
	\caption{Accuracy (\%) comparison on baselines over 10 runs with random initialization. The best result with each dataset are in bold. For our models, the upper results represent the best performance and the lower are the average performance among 10 runs.}
	\label{tab2}
\end{table}

\paragraph{Training Details.}
We utilize Stanford Parser \cite{manning2014stanford} for constituency and dependency parsing. As for our HGCN model, we utilize 300-d GloVe vectors \cite{pennington2014glove} to initialize word embeddings. Then the embeddings are fed into the BiLSTM model, whose hidden size is set to 100-d. The size of dependency relation embeddings are set to 30, while constituency embeddings are set to 100. We use the Adam optimizer with the learning rate of 0.01. HGCN is trained for 100 epochs with batch size 32. The regularization coefficients $\lambda$ is set to 1e-4 and co-efficiency $\gamma$ is set to 3e-2. As for HGCN+BERT, we use the fine-tuned BERT with officially released pre-trained BERT parameters.

\subsection{Results and Analysis}
\begin{figure}[t]
	\centering
	\includegraphics[width=\columnwidth]{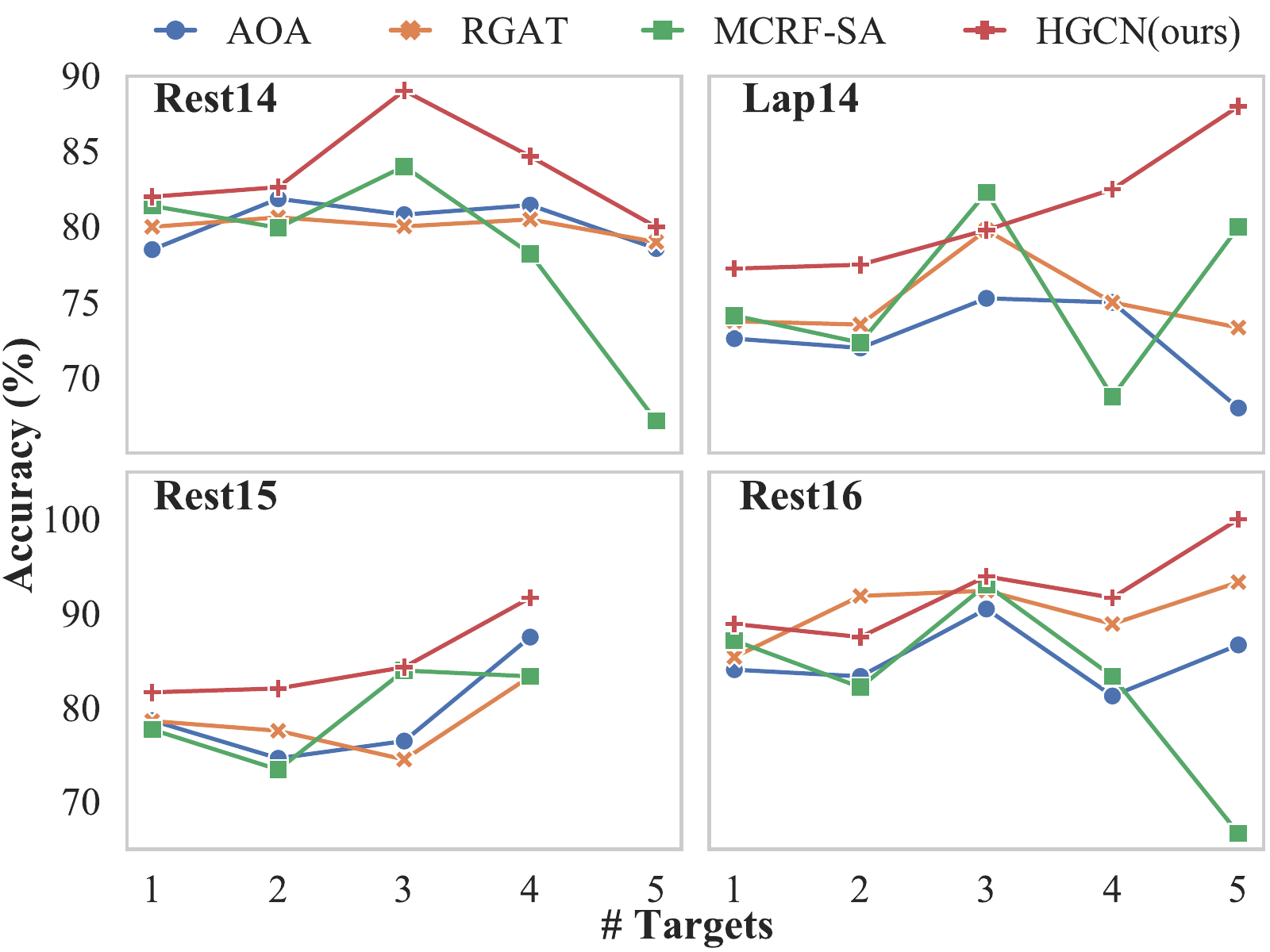} % Reduce the figure size so that it is slightly narrower than the column. Don't use precise values for figure width.This setup will avoid overfull boxes.
	\caption{Accuracy versus the number of targets (\# Targets).}
	\label{fig3}
\end{figure}

\paragraph{Main Performances.}
The overall results are shown in Table \ref{tab2}, from which several observations can be derived. First, the HGCN model consistently outperforms all compared baselines on different datasets. Second, compared with attention-based models, our model is significantly better since it focuses directly on the text regions that are semantically associated with the corresponding target. Therefore, it can eliminate noises introduced by the attention mechanism. Third, the performance of HGCN is also better than syntax-based models. It demonstrates that utilizing constituency and dependency information simultaneously enables the model to capture more structural syntactic information as a way to enhance the representations and HGCN can also avoid multi-hop between target and unrelated opinion words in dependency parsing. As for span-based models, our HGCN is able to model more complete text region compared to them and ensures semantic integrity, which allows our model to have a more robust semantic basis for determining the sentiment polarity. In addition, after we integrate our model with BERT, it obtains further improvement and reaches a new state-of-the-art. In a conclusion, these results can illustrate the effectiveness of our HGCN for capturing structural syntactic information in aspect-level sentiment analysis.

\paragraph{Effect of Multiple Aspects.}

In the datasets, there exists the scenario where a sentence may have multiple targets. Therefore, models may be misled by the opinion words of irrelevant targets and determine the sentiment polarity incorrectly. We conduct experiments on three typical models from different categories mentioned above. As shown in Figure \ref{fig3} (In Rest15, there are no sentences with 5 targets), syntax-based method (RGAT) outperforms attention-based method (AOA) in most cases. It indicates that establishing dependencies between words can partly avoid noisy from unrelated opinion words. As for span-based method (MCRF-SA), it can be seen that the performances tend to fluctuate when the number of targets is more than 3. This is due to the informal expressions and complicated structure of opinion words in sentences. Moreover, it is obvious that our HGCN achieves better performances on four datasets with different number of targets. This confirms the previously established viewpoint that the \emph{Scope} can efficiently delineate the text region of each target so as to alleviate the distraction problem caused by opinion words from irrelevant targets. In addition, it also demonstrates that our HGCN is able to learn more comprehensive syntactic information for reaching robust results.

\begin{table}[t]	
	\centering
	\begin{tabular}{p{2.5cm}<{\centering}cccc}
		\toprule
	    \textbf{Model} & \textbf{Rest14} & \textbf{Lap14} & \textbf{Rest15} & \textbf{Rest16} \cr
		\midrule
		%\hline \rule[0pt]{0pt}{10pt}
		CGCN                & 83.21 & 76.30 & 81.38 & 88.59\cr
		DGCN                & 82.77 & 76.75 & 80.48 & 89.21\cr \hdashline
		\rule[0pt]{0pt}{10pt}CGCN w/o \emph{CRF} 
		                    & 83.13 & 75.68 & 80.64 & 87.81\cr
		DGCN w/o \emph{CRF} & 82.32 & 76.60 & 80.25 & 88.43\cr
		HGCN w/o \emph{CRF} & 83.48 & 76.76 & 80.98 & 88.75\cr \hdashline
		\rule[0pt]{0pt}{10pt}HGCN &\textbf{84.09}&\textbf{78.64}&\textbf{82.66}&\textbf{89.84} \cr
		\bottomrule
	\end{tabular}
	\caption{Experimental results (\%) of ablation study.}
	\label{tab4}
\end{table}

\paragraph{Investigation on the Mask Matrix.}

\begin{figure}[t]
	\centering
	\includegraphics[width=0.85\columnwidth]{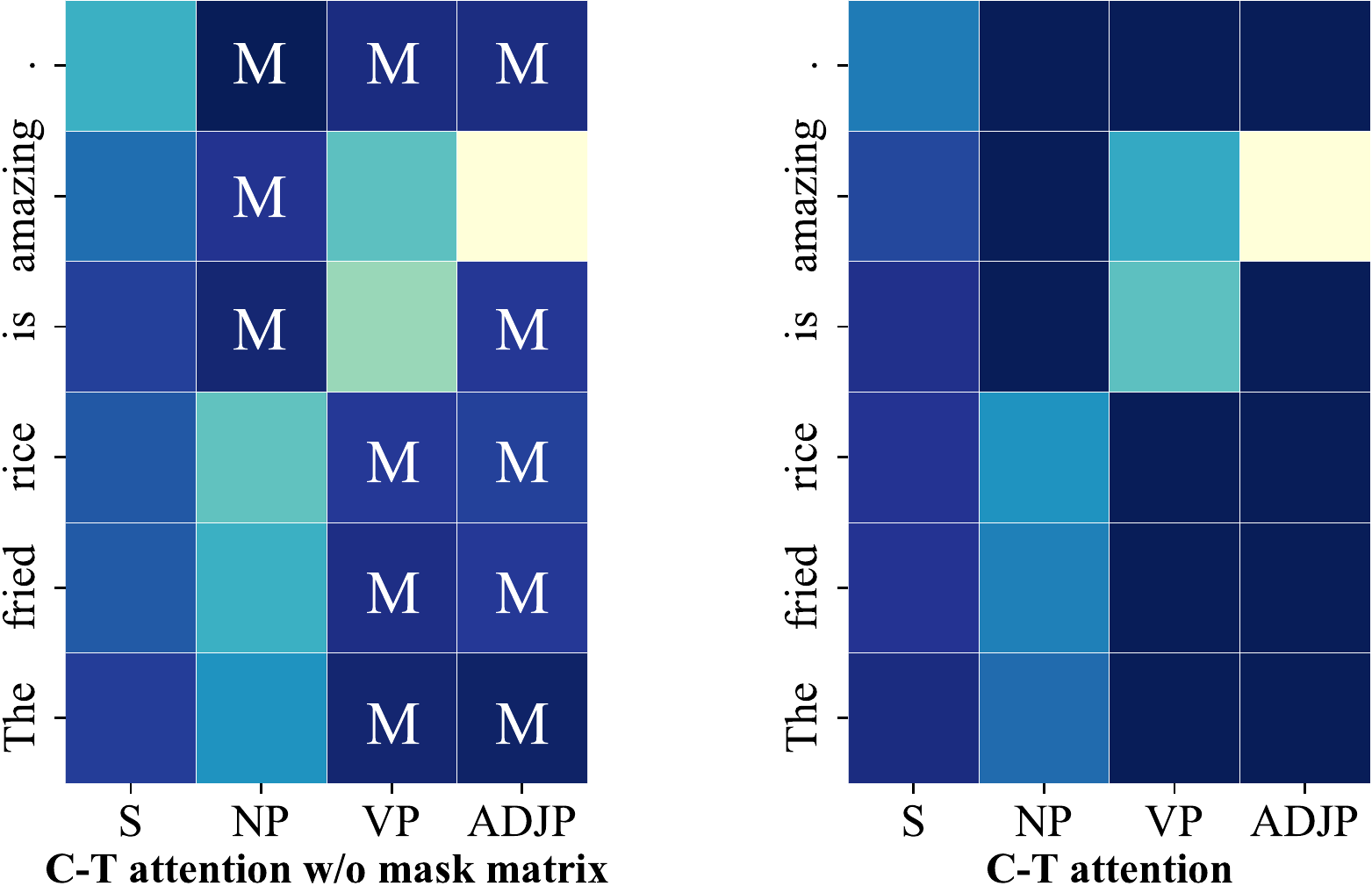} 
	\caption{The C-T attention score matrix of CGCN.}
	\label{fig4}
\end{figure}

In order to investigate the necessity of the mask matrix on assigning the information back to each word from constituents, we visualize the C-T attention score matrix of the CGCN w/o mask (i.e., the mask matrix $M$ in Equation \ref{eq2} is dropped) and intact CGCN. We take the sample shown in Figure \ref{fig2} with its constituency tree. As shown in Figure \ref{fig4}, constituent is able to assign more information to the words that form it based on the attention score matrix constructed by C-T attention mechanism. However, the C-T attention may cause the problem that constituents are forced to inject information back to the words that are not composing them. Though only little information can be assigned to these words, it may introduce interference into CGCN. The mask matrix is utilized to constrain the assignment policy of C-T attention for tackling the problem. Compared with CGCN w/o mask and intact CGCN in Figure \ref{fig4} (M denotes the part that should be masked in C-T attention), the attention score matrix produced by intact CGCN is relatively sparse. For example, only ``amazing’’ can be assigned the information from constituent ``ADJP’’. Therefore, the mask matrix is a necessary component in our proposed CGCN.

\paragraph{Ablation.}

To examine the level of benefits on each component in our model, we conduct extensive ablation studies. The results of accuracy are shown in Table \ref{tab4}. The first observation is that CGCN outperforms DGCN on Rest14 and Rest15, while it fails to act as well as DGCN on the Lap14 and Rest16. Moreover, it is beneficial to incorporate CGCN and DGCN for achieving better performances. This indicates that aggregate syntactic information from two complementary sides of syntactic relations can improve the robustness of the model. More importantly, HGCN w/o CRF means that we remove the subtask of \emph{Scope} modeling so that HGCN cannot utilizing external \emph{Scope} data. However, our HGCN w/o CRF still outperforms the majority of the baselines shown in Table \ref{tab2}. It proves that our motivation of synthesizing constituency tree and dependency tree can learn more comprehensive syntactic information. In addition, the joint method (i.e., our HGCN) is better than the HGCN w/o CRF and achieves the best performance. This confirms that the joint scheme of \emph{Scope} modeling and sentiment polarity determination is effective.

\section{Related Work}

Traditional sentiment analysis works are based on sentence-level or document-level while aspect-level sentiment analysis is a more fine-grained sentiment task.  Therefore, it has received extensive research attentions in recent years.

In general, opinion words are not particularly far from the target term. Thus among neural networks methods, most recent works utilize various attention-based models \cite{wang2016attention,huang2018aspect,fan2018multi} to discriminate sentiment polarity via words nearby the target term implicitly.

Another trends in recent researches are concerned with leveraging dependency parse tree to establish the connection between target terms and opinion words \cite{tang2020dependency}. Typically, GCN is utilized to obtain aspect-oriented features from syntactic trees \cite{zhang2019aspect,sun2019aspect,zhang2020convolution,rui2021dual,liang2020jointly,wang2020relational,tang2020dependency} since it has the ability on addressing the graph structure representation.

Some other efforts try to extract snippet of opinion words directly from the sentence \cite{wang2018learning,xu2020aspect}. These works focus on the explicit extraction of target-opinion pairs to determine sentiment polarity.

In order to obtain constituent representations, we thus propose the constituency GCN model and constituent-token attention in this work for incorporating constituency parse tree  inspired by the graph-based models used in semantic role labeling (SRL) task \cite{marcheggiani2020graph}.

\section{Conclusion}

In this paper, we provide a novel perspective that captures relations between targets and sentiments following the grammatical hierarchy of phrase-clause-sentence structure, aiming at filtering out noisy opinion words from irrelevant targets. Therefore, we introduce the definition of \emph{Scope}, which is a structural text region related to the target term. To learn comprehensive syntactic relations for \emph{Scope}, we propose a hybrid graph convolutional network to synthesize information from constituency tree and dependency tree. Experiments on extensive datasets demonstrate that our model outperforms the baselines and achieves the state-of-the-art performance.

\bibliographystyle{named}
\bibliography{lss4absa}

\appendix

\section{Datasets and Annotation Procedures}\label{app1}

\begin{figure}[t]
	\centering
	\includegraphics[width=\columnwidth]{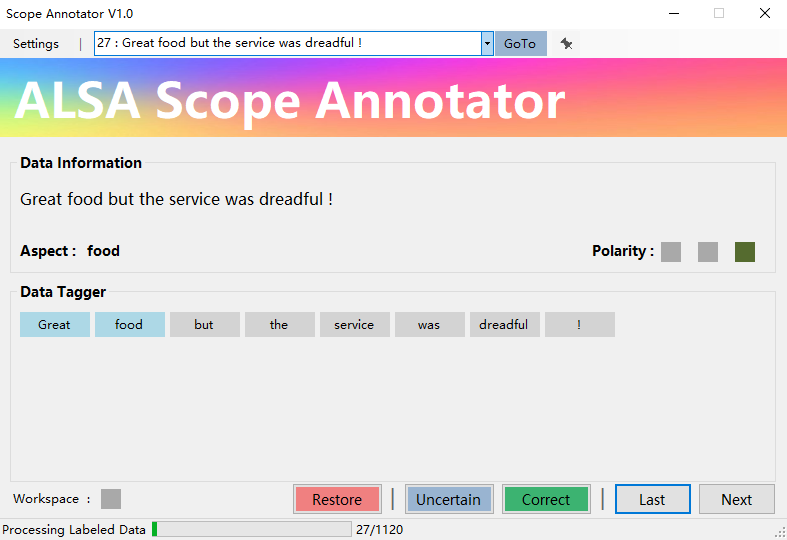} 
	\caption{The interface of our semi-automated \emph{Scope} annotation tool.}
	\label{fig5}
\end{figure}

Consider the sample sentence ``Great food but the service was dreadful!'' with `` \emph{food}'' and ``\emph{service}’’ as targets. The \emph{Scope} of target ``\emph{food}'' can be ``\emph{Great food}'' under definition of the \emph{Scope}. As for the BIO-tags in this case, ``\emph{Great}'' is marked as ``B'' and ``\emph{food}'' is marked as ``I'' while other tokens in the sentence are marked as ``O’’. To simplify the annotation procedures, we develop a semi-automated annotation tool as shown in Figure \ref{fig5}. Our tool can automatically complete the preliminary annotation based on syntax and designed rules. The auto-labeled \emph{Scope} is set as blue blocks in Figure \ref{fig5}. Therefore, the annotators do not need to annotate all samples and only make fine refinements on the inappropriate pre-annotation results. According to our statistics, only 26.4\% of the samples require manual adjustments. This demonstrates that \emph{Scope} annotation procedures are easily accessible and we will open the annotation tool.

\section{Case Study}\label{app2}

In this section, we investigate the behaviour of HGCN, CDT and RGAT on case examples. As we can see from the first example in Table \ref{tab4}, the CDT model is mistakenly identified since it is affected by an incorrect connection in the dependency tree where the edges connect ``\emph{service}'' and ``\emph{great}'' with 2-hops. Due to the gate mechanism, RGAT can filter out the noisy opinion word ``\emph{great}'' by utilizing the edge labels, thus responding correctly. Our HGCN can exploit region information provided by the structural \emph{Scope} to ensure the correctness of the determination. As for the second example, our model reasonably generates the \emph{Scope} related to the target term ``\emph{location}'' and identifies the sentiment polarity correctly. However, due to the edge between target term and ``\emph{treasure}'' in dependency tree, RGAT and CDT are both misled. The third example shows that RGAT and CDT are all confused by another target ``\emph{blueberry}'' in the sentence with its opinion word ``\emph{wonderful}''. By comparison, our model is able to determine sentiment polarity correctly since it delineates the exact context of each target so as to filter out noisy opinion words from irrelevant targets. These examples all illustrate the effectiveness of our \emph{Scope}.

\section{Error Analysis}\label{app3}

\begin{table}[t]	
	\centering
	\begin{tabular}{p{4.3cm}|p{0.8 cm}<\centering p{0.8cm}<\centering p{0.8cm}<\centering}
		\toprule
		Sentence&HGCN&CDT&RGAT\cr 
		\midrule
		%\hline
		\multicolumn{1}{m{4.3cm}|}{1. \textcolor{blue}{Great \textcolor{red}{food}} but \textcolor{blue}{the }\textcolor{red}{service} \textcolor{blue}{was dreadful}!}
		&P$_{\text{\Checkmark}}$N$_{\text{\Checkmark}}$&P$_{\text{\Checkmark}}$P$_{\text{\XSolidBrush}}$&P$_{\text{\Checkmark}}$N$_{\text{\Checkmark}}$ \cr
		
		\multicolumn{1}{m{4.3cm}|}{2. Unfortunately, unless you live in the neighborhood, \textcolor{blue}{it's not in a convenient \textcolor{red}{location}} but is more like a hidden treasure.}    
		&N$_{\text{\Checkmark}}$&P$_{\text{\XSolidBrush}}$&P$_{\text{\XSolidBrush}}$ \cr
		
		\multicolumn{1}{m{4.3cm}|}{3. \textcolor{blue}{My friend had a }\textcolor{red}{burger	} and \textcolor{blue}{i had these wonderful }\textcolor{red}{blueberry pancakes}.}      
		&O$_{\text{\Checkmark}}$P$_{\text{\Checkmark}}$&P$_{\text{\XSolidBrush}}$P$_{\text{\Checkmark}}$&P$_{\text{\XSolidBrush}}$P$_{\text{\Checkmark}}$ \cr
		\bottomrule
	\end{tabular}
	\caption{The words highlighted in red and blue denote the given targets and their \emph{Scope} generated by HGCN. The notations P, N and O represent positive, negative and neutral sentiment, respectively.}
	\label{tab4}
\end{table}

To analyze the limitation of our model, we trace back the error cases in the test sets, and identify three categories of reasons: short-sighted error, interruption error and representation error. The short-sighted error comes from the reason that the \emph{Scope} fails to attend implicit transitive descriptions related to the target term. For example, the sentence ``For someone who used to hate Indian food, Baluchi's has changed my mind.'' with its target term ``\emph{Indian food}'', our \emph{Scope} ignores the latter part of the sentence which contains the semantic transitions. Thus our model considers this to be an expression of negative sentiment polarity resulting in misidentification. As for the interruption error, some grammatical connection of the target term in the sentence (e.g., parenthetical text) can be interruptive, which may interfere with the constituency parsing and the modeling of \emph{Scope}. Consequently, the \emph{Scope} selection will suffer from missing essential information or ill-structured problems, like separate segments. Most of the other errors can be summarized as representation error. Though the model can correctly delineate the \emph{Scope}, it is still unable to determine the sentiment polarity accurately, especially for the expression of negation and modality, which remains challenging for neural networks.

\end{document}